%% file: main.tex
\documentclass[journal]{IEEEtran}

\usepackage[utf8]{inputenc}
\usepackage[T1]{fontenc}
\usepackage{graphicx}
\usepackage{capt-of}
\usepackage{booktabs}
\usepackage{multirow}
\usepackage{amsmath,amssymb,amsthm}
\usepackage{hyperref}
\usepackage[capitalise]{cleveref}
\usepackage{array}
\usepackage{xspace}
\usepackage{pifont}

\newsavebox{\figframebox}

\newcommand{\tracehead}[2]{\vspace{2pt}\noindent{\footnotesize\textbf{#1}\hfill\textbf{#2}}\par\vspace{1pt}}
\newcommand{\tracesetup}{\leftskip=0pt\rightskip=0pt plus 3.5em\parfillskip=0pt plus 1fil}
\newcommand{\tracebody}[1]{{\scriptsize\tracesetup\noindent #1\par}}

\newcommand{\ours}{Document Multi-hop Implicit Derivation \& Extraction}
\newcommand{\oursshort}{DocMIDE\xspace}

\title{\oursshort : Learning Multi-Hop Implicit Derivation in Visually Rich Documents}

\author{Jeremy~Cerwin~Wang, Wai~Kit~Wong, and Jeff Kai~Tai~Tang
\thanks{J. C. Wang and W. K. Wong are with VX Real Limited (e-mail: yuanpeng.wang@vxreal.com; waikit.wong@vxreal.com). W. K. Wong is also with the Hang Seng University of Hong Kong (e-mail: wongwk@hsu.edu.hk). Jeff K. T. Tang is with The Hong Kong Polytechnic University (e-mail: kai-tai-jeff.tang@polyu.edu.hk).}}

\begin{document}

\maketitle

\begin{abstract}
\input{sections/abstract}
\end{abstract}


\input{sections/introduction}
\input{sections/related}
\input{sections/problem}
\input{sections/naive}
\input{sections/method}
\input{sections/experiments/experiments}

\input{sections/limitations}

\newpage

\bibliographystyle{IEEEtran}
\bibliography{bib/references}

\newpage
\appendices
\input{sections/appendix}

\end{document}

%% file: sections/abstract.tex
Real-world document processing systems rely on rigid, predefined schemas, yet critical target fields often lack direct visual counterparts on the page. Extracting these implicit values requires multi-hop derivation, such as aggregating sub-categories or reasoning over visual marks. While existing methods handle explicit text spans or simple implicit queries, they fail at multi-hop visual reasoning even after standard fine-tuning: models retrieve incorrect visual evidence, or retrieve it correctly and then skip the intermediate steps of the derivation. To address this, we introduce \oursshort, a fine-tuning framework that trains compact vision-language models to retrieve visual evidence explicitly before deriving an answer. \oursshort constrains generation to a plan--retrieve--derive structure and optimizes it with Group Relative Policy Optimization under a four-component, rule-based reward that scores output format, the retrieved evidence block, every intermediate derivation step, and the final value against a verified reference trace. On a 4{,}151-pair implicit extraction benchmark, \oursshort raises accuracy from 70.8\% to 95.9\% on Qwen3.5-4B from only a small set of annotated examples, and transfers to a second backbone architecture. Supervised demonstrations alone do not close this gap at any budget we tested; rewarding the intermediate steps is what does.

%% file: sections/introduction.tex
\section{Introduction}
\label{sec:intro}

\begin{figure*}[!t]
\centering
\includegraphics[width=\linewidth]{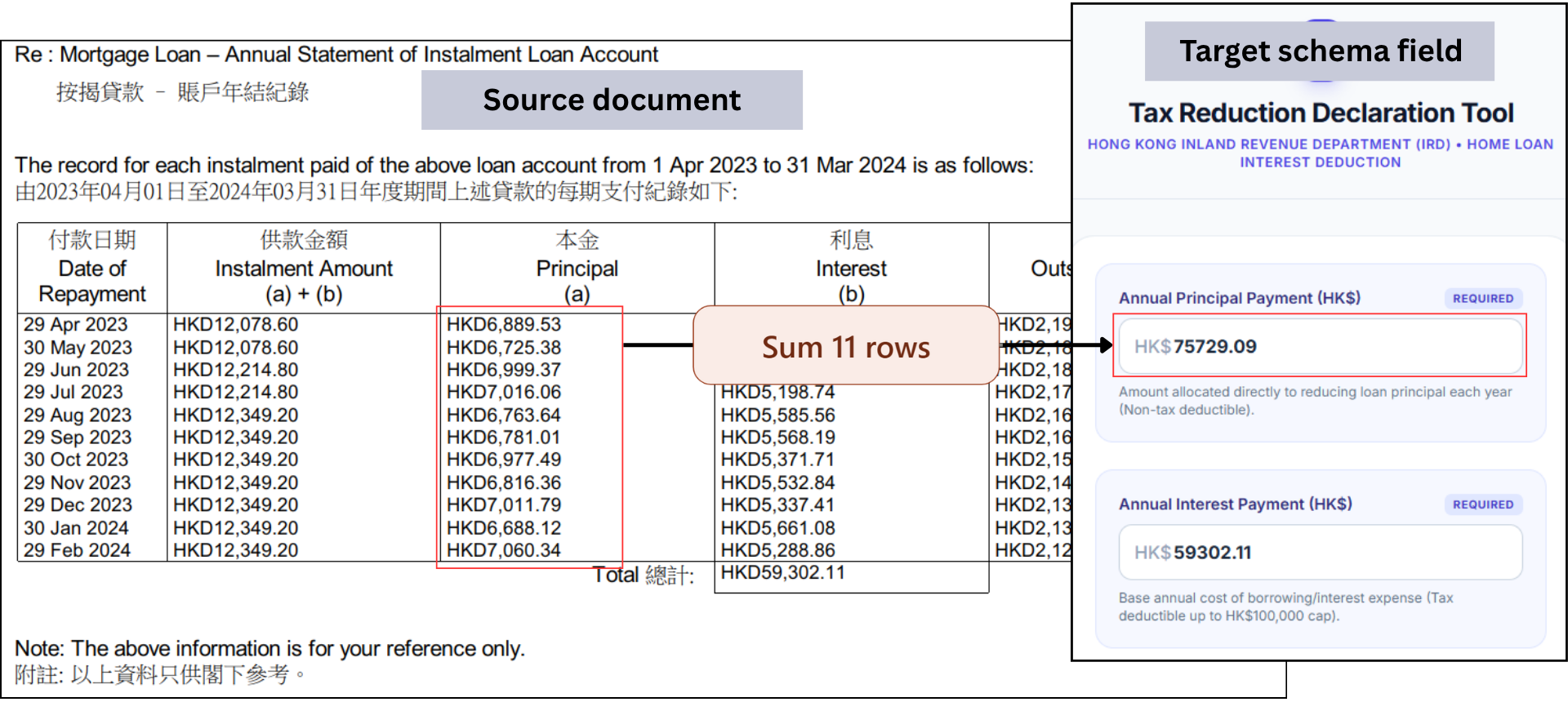}
\caption{Sample document where ``annual principal'' must be derived by summing eleven rows, as the only printed total belongs to an adjacent distractor column (Interest).}
\label{fig:intro-example}
\end{figure*}

\IEEEPARstart{A}{} growing class of AI-powered data platforms extracts structured data with predefined schemas from visually rich documents (VRDs).
However, many critical target fields lack direct visual counterparts on the page.
Extracting these implicit values requires \emph{multi-hop implicit derivation}---chaining evidence scattered across the page through selective category aggregation, multi-operand arithmetic, or interpretation of non-textual visual marks.
Figure~\ref{fig:intro-example} illustrates this challenge in a scanned mortgage statement where the total principal paid is unprinted.
Deriving the field requires summing eleven distinct rows across one column while ignoring an adjacent printed total that serves as a plausible distractor.
Other implicit fields rely entirely on non-textual indicators, such as verifying official stamps or circled form selections.

Addressing this structural gap is critical due to the uncontrolled nature of document ingestion in enterprise workflows.
Platforms operating in domains like ESG compliance, multi-vendor procurement, or job applicant processing must enforce standardized reporting schemas across downstream databases.
However, incoming documents originate from thousands of independent third parties, each using proprietary templates and reporting styles.
For instance, our deployment of an enterprise ESG platform enforcing emissions metrics frequently receives utility bills displaying raw monthly line items rather than annual aggregates, or equipment certificates where compliance is indicated solely by a handwritten checkmark or stamp.
Because the platform cannot force external issuers to alter their layouts, extraction systems cannot rely on simple key-value matching; they must perform multi-hop implicit derivation over whatever visual layout the issuer provides.

To process such heterogeneous layouts, document processing architectures are rapidly shifting from legacy cascades that pair optical character recognition (OCR) with a large language model (LLM) to end-to-end vision-language models (VLMs)~\cite{zhang2024multimodal,lu2024mathvista,rose2023visual}.
While legacy cascaded pipelines\footnote{\raggedright See, e.g., elDoc, ``How to Extract Data from Invoices Using GenAI (OCR + LLM + CV + RAG),'' Nov.\ 2025, \url{https://eldoc.online/blog/how-to-extract-data-from-invoices-using-genai/} (accessed Aug.\ 2026).} strip visual geometry and discard non-textual marks like arrows or circled values during transcription, native VLMs retain full visual context.
However, off-the-shelf VLMs still fall short of industry extraction standards on multi-hop implicit tasks.
Because they are optimized for general multimodal chat rather than schema-constrained derivation, untuned VLMs frequently hallucinate, retrieve misattributed visual evidence, or commit errors during multi-step field derivation.
Worse, standard fine-tuning---even when trained on teacher-filtered reasoning traces from frontier VLMs---fails to bridge this gap.
Supervised fine-tuning (SFT) aligns the shape of a reasoning trace without anchoring it to the page, so the tuned model still commits evidence misassociation and still fabricates intermediate values at inference time (Section~\ref{sec:naive}).


To bridge the gap, we introduce \oursshort (\ours), a lightweight fine-tuning framework that trains VLMs to perform grounded multi-hop implicit derivation.
Rather than predicting field values in a single ungrounded forward pass, \oursshort enforces a structured three-step generation pattern:
(1) an initial reasoning phase that plans required evidence,
(2) an explicit retrieval phase that enumerates raw visual evidence tokens and observations prior to calculation, and
(3) a final reasoning phase that computes the value and emits the answer.

To train models to reliably execute this pattern, \oursshort leverages Group Relative Policy Optimization (GRPO)~\cite{shao2024deepseekmath} guided by a novel four-component, rule-based reward function:
(1) \textbf{Format Coherence:} enforces structural discipline;
(2) \textbf{Grounding Quality:} rewards high recall and precision when identifying raw evidence spans from the document;
(3) \textbf{Stepwise Reasoning:} verifies intermediate derivation steps against the model's retrieved evidence; and
(4) \textbf{Answer Accuracy:} evaluates the correctness of the final output value.

Extensive empirical evaluation shows that \oursshort increases held-out extraction accuracy from 70.8\% to 95.9\% on Qwen3.5-4B~\cite{qwen2026qwen35} and from 13.5\% to 82.6\% on InternVL3.5-4B~\cite{wang2025internvl35}, outperforming state-of-the-art open-weight VLMs~\cite{yu2026minicpm,chen2026eagle,dong2026qianfanocr} and OCR-based methods~\cite{duan2026glm,wu2026firered,wei2026deepseek,cui2025paddleocr3,yin2026unlimited,bhattacharyya2025information} (Section~\ref{sec:experiments}).

%% file: sections/related.tex
\section{Related Work}
\label{sec:related}

\vspace{4pt}\noindent\textbf{Implicit inference in general vision and NLP.}
Outside of document processing, implicit inference has been explored in text-based question answering~\cite{liu2022disentangled}, video relational reasoning~\cite{swetha2026vrr}, and structured reasoning traces for implicit-knowledge visual question answering (VQA)~\cite{wen2026star} (StaR-KVQA).
Vision-R1~\cite{huang2026vision} further applies long chain-of-thought (CoT) reasoning to vision-language models, representing the closest published system to our training setup.
While Vision-R1 and StaR-KVQA share conceptual similarities with our approach, these methods generally focus on open-ended question answering over general images or video rather than structured document environments.
As a result, they rely on retrieving external commonsense knowledge or interpreting implicit motives to output free-form text.
In contrast, document-implicit fields require extracting unprinted quantities or marks derivable strictly by chaining visual evidence across a page to satisfy a rigid target schema.
Existing extraction frameworks struggle with these multi-hop implicit derivations, frequently failing when forced to decompose complex multi-step computations without explicit intermediate evidence grounding.

\vspace{4pt}\noindent\textbf{Extraction architectures.}
Standard document extraction typically falls into two paradigms: traditional OCR pipelines and end-to-end VLMs.
Traditional OCR pipelines employ an OCR engine (GLM-OCR~\cite{duan2026glm}, FireRed-OCR~\cite{wu2026firered}, DeepSeek-OCR 2~\cite{wei2026deepseek}, PaddleOCR~\cite{cui2025paddleocr3}, Unlimited-OCR~\cite{yin2026unlimited}, BLOCKIE~\cite{bhattacharyya2025information}) to convert document pages into flat text before passing the output to a downstream LLM for field extraction.
However, converting visual layouts into linear text causes an irreversible loss of spatial relationships and structural semantics: non-textual indicators like circled choices or directional arrows are completely lost during transcription, depriving downstream LLMs of critical visual evidence.
While OCR systems continue to advance, they share the same transcribe-then-parse ceiling—untranscribed marks cannot be recovered downstream.
Conversely, end-to-end VLMs~\cite{zhang2024multimodal,lu2024mathvista,rose2023visual,yu2026minicpm} preserve raw visual inputs.
However, state-of-the-art general-purpose VLMs (Qwen3.5-4B~\cite{qwen2026qwen35}, InternVL3.5-4B~\cite{wang2025internvl35}) focus on broad multimodal tasks, whereas specialized document VLMs (Eagle2.5-8B~\cite{chen2026eagle}, Qianfan-OCR~\cite{dong2026qianfanocr}) are tailored for direct visual extraction rather than schema-constrained, multi-hop implicit derivation.

\vspace{4pt}\noindent\textbf{Benchmarks.}
Existing document benchmarks predominantly focus on explicit, span-extractive fields.
Datasets like DocILE~\cite{simsa2023docile} restrict annotations to explicit character spans, while form-key-value linking datasets (XFUND~\cite{xu2022xfund}, SRFUND~\cite{ma2024srfund}, EPHOIE~\cite{wang2021ephoie}) rely strictly on explicit visual key anchors.
Region-level grounding benchmarks (OCRBench v2~\cite{fu2026ocrbench}, CC-OCR~\cite{yang2025ccocr}, MMDocBench~\cite{zhu2026mmdocbench}) and financial QA suites (TAT-DQA~\cite{zhu2022towards}, FinQA~\cite{chen2021finqa}, DocMath-Eval~\cite{zhao2024docmath}) focus on single-step localization or free-form numerical QA.
Similarly, general document parsing (OmniDocBench~\cite{ouyang2025omnidocbench}), document question answering (DocVQA~\cite{mathew2021docvqa}), and infographic reasoning benchmarks (ChartQAPro~\cite{masry2025chartqapro}) assess broad layout transcription, open-ended visual QA, or chart analysis without fixed-schema constraints.
Finally, while datasets like UNIKIE-Bench~\cite{ji2026unikie} capture diverse real-world administrative, commercial, advertising, and accommodation documents, they do not systematically isolate multi-step visual derivations from explicit extractions.
Consequently, none of these existing benchmarks are equipped to evaluate schema-constrained, multi-hop implicit derivation.

\vspace{4pt}\noindent\textbf{Policy optimization.}
To reinforce multi-step reasoning without the overhead of an explicit critic network, GRPO~\cite{shao2024deepseekmath} has emerged as a widely adopted reinforcement learning framework.
While GRPO was originally designed to enhance mathematical and logical reasoning in language models, recent efforts have effectively extended it to multimodal settings, such as visual reasoning~\cite{huang2026vision, cao2025ground, masry2025bigcharts}.
Building on this common practice of using GRPO to cultivate structured reasoning capabilities, we fine-tune our models via Low-Rank Adaptation (LoRA)~\cite{hu2021lora} guided by a reward function tailored for grounded evidence retrieval and schema-constrained derivation.

%% file: sections/problem.tex
\section{Problem Formulation: Multi-Hop Implicit Derivation}
\label{sec:problem}

We define document field extraction as a function $f: (I, S) \rightarrow V$, where $I$ is a document image, $S = \{s_1, \ldots, s_n\}$ is a predefined schema of target fields, and $V = \{v_1, \ldots, v_n\}$ is the set of extracted values. 
Let $\text{OCR}(I) = \{x_1, x_2, \ldots, x_m\}$ represent the set of discrete textual primitives transcribed from $I$ (alongside non-textual visual marks present on the page).
A field $s_i \in S$ is \emph{explicit} if its value appears directly as a primitive span ($v_i \in \text{OCR}(I)$).
Conversely, $s_i$ is \emph{implicit} if its value is not directly present in $\text{OCR}(I)$ ($v_i \notin \text{OCR}(I)$) and must instead be computed through a directed sequence of elementary operations.

Let $\mathcal{O}$ denote a bounded set of \emph{atomic primitive operators} (e.g., binary arithmetic $\text{add}(u, w)$, relational category predicate matching $\text{match}(u, c)$, or unary visual evaluation $\text{is\_checked}(u)$), where each operator $\phi \in \mathcal{O}$ acts on at most two operands to yield an intermediate state:
\begin{equation*}
z = \phi(u, w), \quad \text{where } \phi \in \mathcal{O}.
\end{equation*}
Because every operator $\phi \in \mathcal{O}$ is strictly atomic, complex field extractions cannot be collapsed into a single operational mapping.

We define a $k$-hop derivation chain for a target field $s_i$ as a sequence of intermediate states $(z_1, z_2, \ldots, z_k)$ constructed as follows:
\begin{align*}
z_1 &= \phi_1(u_1, w_1) \; \text{where } u_1, w_1 \in \text{OCR}(I), \\
z_2 &= \phi_2(u_2, w_2) \; \text{where } u_2, w_2 \in \text{OCR}(I) \cup \{z_1\}, \\
&\;\;\vdots \\
z_k &= \phi_k(u_k, w_k) \; \text{where } u_k, w_k \in \text{OCR}(I) \cup \{z_1, \dots, z_{k-1}\}.
\end{align*}
The field extraction task is formally classified as \emph{multi-hop implicit derivation} if and only if the minimal valid derivation chain yielding the final value $v_i = z_k$ requires a depth of $k \ge 2$ operational steps.

Under this formulation, the model is given only the image $I$ and the schema field description $s_i \in S$ at inference time. 
To populate $v_i$, the model must implicitly discover the sequence of operators $(\phi_1, \ldots, \phi_k)$ and select the appropriate primitive operands $(u_j, w_j)$ without explicit step-by-step guidance.

\vspace{4pt}\noindent\textbf{Deployment constraints and efficiency target.} 
In addition to the multi-hop requirement, we specifically target \emph{small vision-language models} (e.g., $\le 4\text{B}$ parameters). 
While modern trillion-parameter API models can perform multi-step execution via brute-force scaling, real-world document processing pipelines require low-latency, privacy-preserving, and cost-effective deployments. 
Accordingly, our task formulation assumes that the extraction model must be fully trainable and runnable on a \emph{single commodity GPU} (e.g., an RTX 4090), making reliable implicit derivation within compact architectures a primary research challenge.

%% file: sections/naive.tex
\section{Does Standard Fine-Tuning Suffice?}
\label{sec:naive}

\begin{figure*}[!t]
\centering
\setlength{\fboxsep}{8pt} 
\setlength{\fboxrule}{0.8pt} 
\fbox{%
\begin{minipage}{\dimexpr\textwidth-2\fboxsep-2\fboxrule\relax}
\scriptsize
\raggedright
\tracehead{Document image}{}\par\vspace{3pt}
\centering
\includegraphics[width=\textwidth]{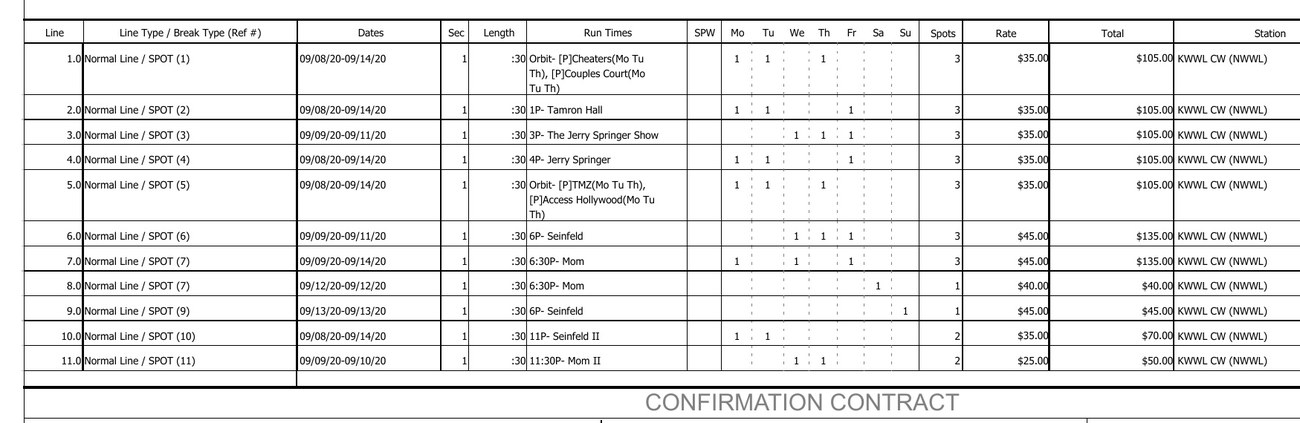}\par\vspace{8pt}
\raggedright
\tracehead{Target field}{Ground Truth: \textbf{1000.00}}\par\vspace{3pt}
\textbf{Key:} \texttt{eleven\_lines\_grand\_total} \quad|\quad \textbf{Description:} \emph{The sum of the Total column across all eleven line items on this invoice.}\par\vspace{8pt}
\tracehead{Target output}{}\par\vspace{2pt}
\tracebody{\texttt{<think>This KWWL confirmation contract for House Majority PAC lists eleven separate line items across a mix of syndicated talk and sitcom programs — Cheaters/Couples Court, Tamron Hall, two separate Jerry Springer lines, TMZ/Access Hollywood, two separate Seinfeld lines, two separate Mom lines, and Seinfeld II — each with its own spot count and rate, so no single pattern lets you skip reading every row. The eleven line Totals, read down the table in order, are 105.00, 105.00, 105.00, 105.00, 105.00, 135.00, 135.00, 40.00, 45.00, 70.00, and 50.00. Adding them one at a time: 105.00 + 105.00 = 210.00, then 210.00 + 105.00 = 315.00, then 315.00 + 105.00 = 420.00, then 420.00 + 105.00 = 525.00, then 525.00 + 135.00 = 660.00, then 660.00 + 135.00 = 795.00, then 795.00 + 40.00 = 835.00, then 835.00 + 45.00 = 880.00, then 880.00 + 70.00 = 950.00, then 950.00 + 50.00 = 1,000.00. Since the prompt asks for the number only, the final answer is 1000.00.</think> 1000.00}}
\end{minipage}%
}
\caption{A training sample under standard fine-tuning. The target output is generated by a frontier VLM and manually verified. Reasoning steps are contained entirely within the \texttt{<think>} tag, with the final extracted field value appended afterward.}
\label{fig:sft-example}
\end{figure*}

A natural question is whether standard fine-tuning on high-quality reasoning traces can resolve multi-hop implicit derivation. 
To construct a strong fine-tuning baseline, we first generate step-by-step reasoning trajectories using frontier closed-source models (e.g., Claude, OpenAI models) and manually inspect both the intermediate reasoning steps and final answers. 
Filtering for strict correctness yields a curated pool of $200$ high-quality exemplars used for initial SFT on open-source vision-language baselines, followed by policy optimization via GRPO.
The baseline reported in our main results draws $50$ of these exemplars, matching the supervision budget given to \oursshort so that the comparison isolates the training scheme; Appendix~\ref{sec:appendix_sft_scale} verifies that fine-tuning on the full pool of $200$ does not change the outcome.
Figure~\ref{fig:sft-example} shows one such exemplar in full.

While this two-phase fine-tuning pipeline aligns models to output structured CoT traces, our evaluation shows that standard fine-tuning alone is insufficient to guarantee reliable multi-hop execution at inference time (detailed in Section~\ref{sec:experiments}). 
Instead, fine-tuned baselines suffer from two compounding failure modes:
\begin{itemize}
    \item \textbf{Evidence Misassociation:} In complex layouts, models frequently pull distractor elements (e.g., values from adjacent columns, wrong line items, or a nearby field holding the same kind of value) into their reasoning chain, leading to ungrounded intermediate steps.
Figure~\ref{fig:fail-evidence} shows one such trace in full.

\begin{figure}[!t]
\centering
\setlength{\fboxsep}{8pt} 
\setlength{\fboxrule}{0.8pt} 
\fbox{%
\begin{minipage}{\dimexpr\linewidth-2\fboxsep-2\fboxrule\relax}
\scriptsize
\raggedright
\tracehead{Document image}{}\par\vspace{3pt}
\centering
\includegraphics[width=\linewidth]{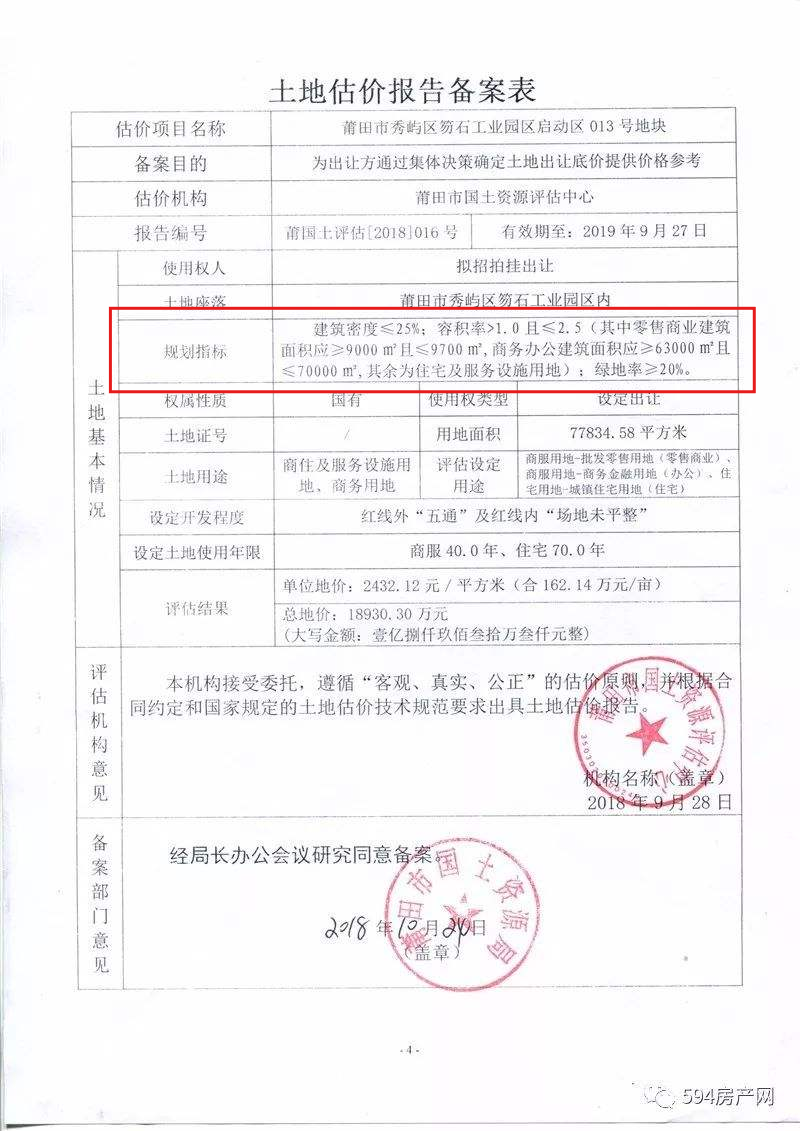}\par\vspace{8pt}
\raggedright
\tracehead{Target field}{Ground Truth: \textbf{7000}}\par\vspace{3pt}
\textbf{Key:} \texttt{office\_area\_range\_width} \quad|\quad \textbf{Description:} \emph{The width of the permitted Office Building Area range stated in the Planning Indicators section, maximum minus minimum, in square metres.}\par\vspace{8pt}
\tracehead{Standard fine-tuning output}{Answered: \textbf{700} \ding{55}}\par\vspace{2pt}
\tracebody{\texttt{<think> The Planning Indicators section states that the Office Building Area must be at least 9,000 m\textsuperscript{2} and at most 9,700 m\textsuperscript{2}. Subtracting the minimum from the maximum: 9,700 - 9,000 = 700, which is the final answer.</think> 700}}
\end{minipage}%
}
\caption{\textbf{An example of evidence misassociation under standard fine-tuning.} The model extracts the wrong operands ($9{,}700$ and $9{,}000$) instead of the correct values ($70{,}000$ and $63{,}000$).}
\label{fig:fail-evidence}
\end{figure}

    \item \textbf{Superficial Mimicry and Step-Skipping:} While SFT teaches models to mimic the superficial formatting of intermediate steps, it fails to enforce step-by-step rigor. Models frequently jump operational steps or hallucinate ungrounded intermediate values rather than systematically executing the derivation chain. Figure~\ref{fig:fail-mimicry} shows one such trace in full.

\begin{figure}[!t]
\centering
\setlength{\fboxsep}{8pt} 
\setlength{\fboxrule}{0.8pt} 
\fbox{%
\begin{minipage}{\dimexpr\linewidth-2\fboxsep-2\fboxrule\relax}
\scriptsize
\raggedright
\tracehead{Document image}{}\par\vspace{3pt}
\centering
\includegraphics[width=\linewidth]{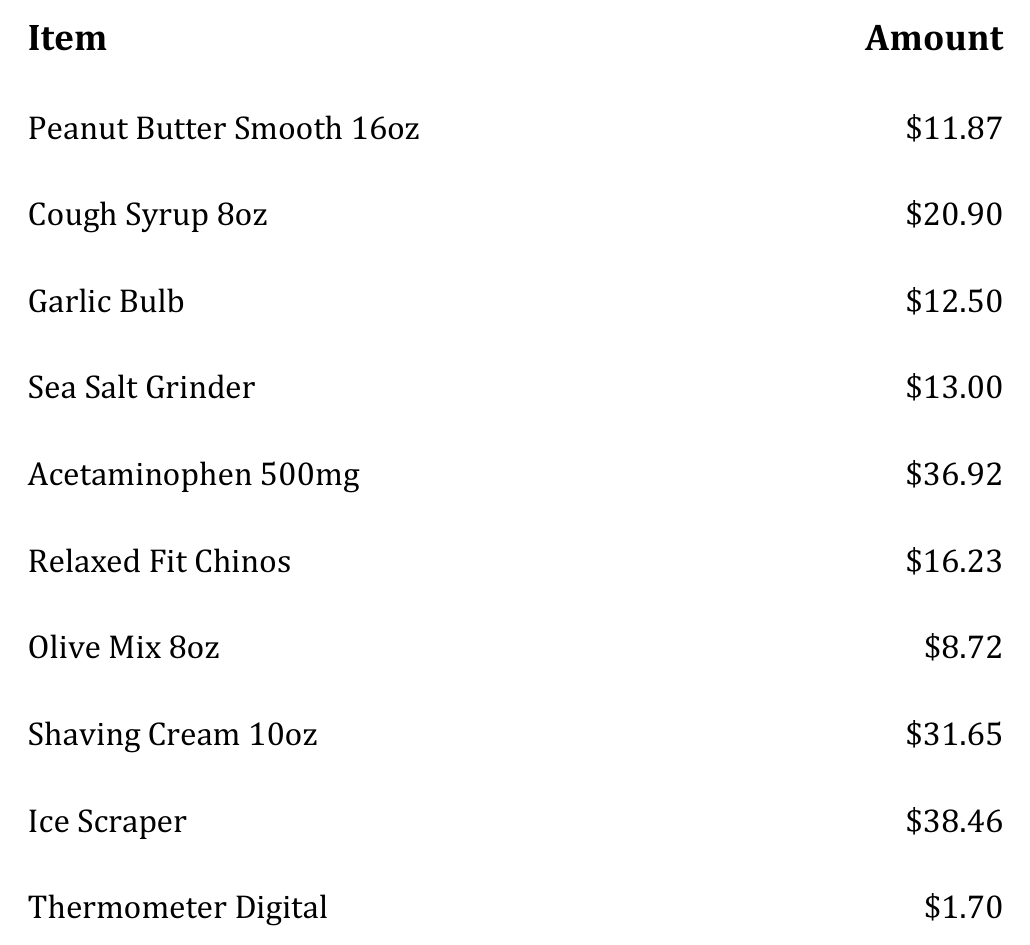}\par\vspace{8pt}
\raggedright
\tracehead{Target field}{Ground Truth: \textbf{191.95}}\par\vspace{3pt}
\textbf{Key:} \texttt{amount\_total} \quad|\quad \textbf{Description:} \emph{The sum of the Amount column across all ten line items on this receipt.}\par\vspace{8pt}
\tracehead{Standard fine-tuning output}{Answered: \textbf{192.95} \ding{55}}\par\vspace{2pt}
\tracebody{\texttt{<think> This receipt from Harvest Moon Market in Tucson lists 10 distinct items, ranging from everyday groceries like Peanut Butter Smooth and Garlic Bulb to household goods like Sea Salt Grinder and Ice Scraper, and even some personal care items like Shaving Cream and Cough Syrup. Each item has its own price, and the final answer is the sum of all 10 amounts: 11.87 + 20.90 + 12.50 + 13.00 + 36.92 + 16.23 + 8.72 + 31.65 + 38.46 + 1.70 = 192.95, which is the final answer as a bare number.</think> 192.95}}
\end{minipage}%
}
\caption{\textbf{An example of superficial mimicry and step-skipping under standard fine-tuning.} The model asserts a single ten-term sum instead of computing step-by-step additions, resulting in an arithmetic error ($192.95$ vs. $191.95$).}
\label{fig:fail-mimicry}
\end{figure}
\end{itemize}

These qualitative failure modes motivate the core design of \oursshort, which moves beyond standard single-pass fine-tuning by explicitly enforcing grounded evidence verification and structured multi-stage execution.

%% file: sections/method.tex
\section{\ours}
\label{sec:framework}

\subsection{Response Structure}
\label{sec:response_structure}

To enforce grounded reasoning and make each derivation stage explicitly verifiable, \oursshort restricts model outputs to a structured three-step generation pattern:
\begin{enumerate}
    \item \textbf{Initial Reasoning Phase:} The model analyzes the schema field requirements and formulates a plan to locate the necessary visual evidence on the page.
    \item \textbf{Explicit Retrieval Phase:} The model explicitly enumerates all raw visual evidence tokens and values retrieved from the document prior to performing any calculation.
    \item \textbf{Final Reasoning Phase:} The model performs step-by-step reasoning over the extracted evidence block to compute intermediate states and emits the final schema answer.
\end{enumerate}

In our implementation, the explicit retrieval phase is visually and syntactically isolated using special system tags (\texttt{<raw>} and \texttt{</raw>}).
Forcing the model to declare raw visual evidence in a dedicated intermediate step serves two critical functional purposes:
\begin{enumerate}
    \item \textbf{Complete Contextual Assembly:} It forces the model to articulate all requisite visual and textual evidence within the context window before execution, ensuring downstream reasoning is conditioned on a complete set of retrieved facts.
    \item \textbf{Explicit Verifiability and Optimization:} Isolating retrieved evidence into a discrete text block renders visual grounding independently verifiable. 
This allows us to construct a targeted reinforcement learning reward that specifically trains the model to maximize evidence retrieval accuracy prior to derivation.
\end{enumerate}

\vspace{4pt}\noindent\textbf{Training Data Construction and Labeling.}
To train the base model to emit this structured generation pattern, each target field $s_i \in S$ in the dataset is formatted as a 4-tuple $(\hat{t}_\text{plan}, \hat{E}, \hat{t}_\text{exec}, \hat{v})$ consisting of:
\begin{enumerate}
    \item \textbf{Planning Trace ($\hat{t}_\text{plan}$):} The initial reasoning text articulating the retrieval goal and target schema constraints for field $s_i$.
    \item \textbf{Grounding Evidence Block ($\hat{E}$):} A structured block containing the extracted ground-truth evidence operands $\hat{E} = \{u_1, u_2, \dots, u_{|\hat{E}|}\} \subseteq \text{OCR}(I)$ from the document image.
    \item \textbf{Stepwise Execution Trace ($\hat{t}_\text{exec}$):} A sequence of explicit intermediate derivations mapping directly to the $k$-hop operational chain:
    \begin{equation*}
        \hat{z_j} = \phi_j(u_j, w_j) \quad \text{for } j = 1, \dots, k
    \end{equation*}    
    \item \textbf{Ground Truth ($\hat{v}$):} The final extracted value $\hat{v} = \hat{z_k}$.
\end{enumerate}
An example training completion is shown in Figure~\ref{fig:our-example}.

\begin{figure}[!t]
\centering
\fbox{%
\begin{minipage}{\dimexpr\linewidth-2\fboxsep-2\fboxrule\relax}
{\scriptsize\tracesetup
\textbf{1. Planning Trace ($t_\text{plan}$):}\par
\texttt{To answer this, I need to identify: line totals.}\par\vspace{6pt}

\textbf{2. Grounding Evidence Block ($\hat{E}$):}\par
\texttt{105.00, 105.00, 105.00, 105.00, 105.00, 135.00, 135.00, 40.00, 45.00, 70.00, 50.00}\par\vspace{6pt}

\textbf{3. Stepwise Execution Trace ($t_\text{exec}$):}\par
\texttt{$\phi_1(105.00, 105.00) = 210.00$, $\phi_2(210.00, 105.00) = 315.00$, ..., $\phi_{10}(950.00, 50.00) = 1000.00$, where $\phi_i(u, w) = u + w$ is an addition operator for all $i \in [1, 10]$.}\par\vspace{6pt}

\textbf{4. Ground Truth ($\hat v$):}\par
\texttt{1000.00}\par}
\end{minipage}%
}
\caption{\oursshort structured training sample for the document prompt in Fig.~\ref{fig:sft-example}. The label explicitly partitions the target generation into four components: planning trace ($t_\text{plan}$), explicit evidence grounding block ($\hat E$), pairwise stepwise execution trace ($t_\text{exec}$) where each operator $\phi_i$ for $i \in [1, 10]$ is addition, and final ground-truth scalar answer $\hat v$.}
\label{fig:our-example}
\end{figure}

\subsection{Training Overview}

Training \oursshort proceeds in two sequential stages: supervised fine-tuning followed by policy optimization.

\textbf{Stage 1: Supervised Fine-Tuning.}
The primary objective of the SFT phase is to cold-start the base model and teach it to strictly adhere to the three-step response structure defined in Section~\ref{sec:response_structure}.

\textbf{Stage 2: Custom GRPO.}
We employ a custom implementation of GRPO~\cite{shao2024deepseekmath} as the primary training phase to optimize the SFT checkpoint policy $\pi_\theta$.
Given an input document prompt $q$, GRPO samples a group of $G$ candidate outputs $\{o_1, o_2, \dots, o_G\}$ from the old policy $\pi_{\theta_{old}}$.
Each generated completion $o_i$ is assigned a scalar reward $R_i = R(q, o_i)$ by a task-specific reward function $R$.
The relative advantage $\hat{A}_i$ for completion $o_i$ is computed by standardizing the rewards within the sampled group:
\begin{equation*}
    \hat{A}_i = \frac{R_i - \text{mean}(\{R_1, \dots, R_G\})}{\text{std}(\{R_1, \dots, R_G\}) + \epsilon}
\end{equation*}
where $\epsilon$ is a small constant to ensure numerical stability.
This relative advantage directly reinforces outputs that outperform group averages while suppressing inferior ones without requiring a separate critic network.

Because optimization depends entirely on relative intra-group advantages $\hat{A}_i$, the critical driver of performance is the design of the reward function $R(q, o_i)$.
The formulation and components of our reward function are detailed in Section~\ref{sec:reward}.

\input{sections/reward/reward.tex}

%% file: sections/reward/reward.tex
\subsection{Reward Design}
\label{sec:reward}

We score each sampled completion using four deterministic, rule-based components—\textbf{Format Coherence} ($R_f$), \textbf{Grounding Quality} ($R_g$), \textbf{Stepwise Reasoning} ($R_s$), and \textbf{Answer Accuracy} ($R_a$)—designed to isolate and penalize distinct failure modes in multi-step visual reasoning without relying on a learned reward model.

\vspace{4pt}\noindent\textbf{Total Composite Reward.}
Let $\hat{y} = (\hat{t}_\text{plan}, \hat{E}, \hat{t}_\text{exec}, \hat{v})$ denote the target ground-truth label corresponding to document query $q$.
The overall scalar reward $R(q, o_i)$ passed to the GRPO objective for a candidate output $o_i$ and document prompt $q$ is a linearly weighted sum of four sub-rewards, where each sub-reward component outputs a bounded score in $[0, 1]$:
\begin{equation*}
\begin{split}
R(q, o_i) = {} & \lambda_f R_f(o_i) + \lambda_g R_g(\hat{y}, o_i) \\
               & {} + \lambda_s R_s(\hat{y}, o_i) + \lambda_a R_a(\hat{y}, o_i)
\end{split}
\end{equation*}
where $\boldsymbol{\lambda} = (\lambda_f, \lambda_g, \lambda_s, \lambda_a)$ are positive system hyperparameters.
By default, we set $\lambda_f = 0.25$, $\lambda_g = 1.0$, $\lambda_s = 0.75$, and $\lambda_a = 2.5$.
Through internal hyperparameter tuning, we found that performance remains robust to minor weight variations.

\vspace{4pt}\noindent\textbf{1. Format Coherence ($R_f$).}
Similar to standard GRPO formatting rewards, $R_f(o_i)$ serves as a structural regularizer.
It encourages the model to adhere to the response structure defined in Section~\ref{sec:response_structure}.
It assigns a fractional score based on the proportion of required structural components present in the output trace.
When correctly formatted, the model output $o_i$ can be uniquely parsed into the 4-tuple $(t_\text{plan}, E, t_\text{exec}, v)$ corresponding to the target structure $(\hat{t}_\text{plan}, \hat{E}, \hat{t}_\text{exec}, \hat{v})$.
Specifically, $t_\text{exec}$ is extracted by applying regular expression pattern matching over structural tags to identify operational steps matching the set of atomic primitive operators $\mathcal{O}$.

\input{sections/reward/ground.tex}

\input{sections/reward/reason.tex}

\vspace{4pt}\noindent\textbf{4. Answer Accuracy ($R_a$).}
The answer accuracy reward $R_a(\hat{y}, o_i)$ provides a strict binary outcome evaluation, assigning $1$ if the extracted final value $v \in o_i$ exactly matches the ground truth $\hat{v} \in \hat{y}$, and $0$ otherwise.

%% file: sections/reward/ground.tex
\vspace{4pt}\noindent\textbf{2. Grounding Quality ($R_g$).}
The grounding reward $R_g(\hat{y}, o_i)$ penalizes evidence misassociation by evaluating the candidate evidence block $E \in o_i$ against the ground-truth evidence block $\hat{E} \in \hat{y}$.
We compute precision and recall over multisets of primitive tokens to ensure $E$ contains the exact target evidence elements.
Furthermore, to verify that the extracted evidence in $E$ is actively consumed during calculation, we evaluate trace utility within $t_\text{exec} \in o_i$.
Specifically, we collect all operands across the executed operations $\{\phi_j\}$ in $t_\text{exec}$ and compute the fraction of elements in $E$ that appear in this set, accounting for multiset element frequencies.
The sub-scores are formally defined as:
\begin{align*}
\mathrm{precision} &= \frac{|F(\hat{E}) \cap F(E)|}{|F(E)|}, \\
\mathrm{recall} &= \frac{|F(\hat{E}) \cap F(E)|}{|F(\hat{E})|}, \\
\mathrm{utility} &= \frac{|F(t_\text{exec}) \cap F(E)|}{|F(\hat{E})|},
\end{align*}
where $F(x)$ extracts the multiset of primitive visual and textual tokens from a block $x$.
The overall grounding score is computed as:
\begin{equation}
R_g(\hat{y}, o_i) = \tfrac{1}{3}\left(\mathrm{precision} + \mathrm{recall} + \mathrm{utility}\right).
\end{equation}
Missing or empty evidence blocks ($|F(E)| = 0$) directly assign $R_g(\hat{y}, o_i) = 0$.

%% file: sections/reward/reason.tex
\vspace{4pt}\noindent\textbf{3. Stepwise Reasoning ($R_s$).}
The stepwise reasoning reward $R_s(\hat{y}, o_i)$ evaluates execution fidelity by measuring ordered derivation accuracy in $t_\text{exec} \in o_i$ relative to target trace $\hat{t}_\text{exec} \in \hat{y}$.
For target trace $\hat{t}_\text{exec} = (\hat{s}_1, \dots, \hat{s}_k)$ of length $k$ and generated trace $t_\text{exec} = (s_1, \dots, s_n)$, we compute the length of their longest common subsequence $\ell_\text{seq} = |\mathrm{LCS}(t_\text{exec}, \hat{t}_\text{exec})|$.
This ensures that generated steps are credited if they match target step tuples while strictly preserving the correct operational ordering.
The normalized stepwise reward is defined as:
\begin{equation}
R_s(\hat{y}, o_i) = \frac{\ell_\text{seq}}{k}.
\end{equation}

However, strict sequence matching is overly rigid for multi-operand associative operations (e.g., summing $N$ quantities), where multiple valid operational orderings exist.
To accommodate this flexibility, we employ a dynamic pattern-matching mechanism for arithmetic chains.
Specifically, we track the set of consumed raw evidence inputs and verified intermediate outputs: the initial step must combine two raw ground-truth values, while each subsequent step $j$ is valid if it operates on the verified result of step $j-1$ and an unconsumed raw value.
Each sequentially valid step matching this accumulator pattern increments the matched step count $\ell_\text{seq}$, rewarding arbitrary yet logically sound summation orderings while penalizing disconnected or hallucinated calculations.

%% file: sections/experiments/experiments.tex
\section{Experiments}
\label{sec:experiments}

In this section, we empirically evaluate the effectiveness of \oursshort.
We first outline our experimental configuration, including benchmark datasets, evaluation protocols, and implementation details.
We then present our primary results comparing \oursshort against state-of-the-art VLMs, traditional OCR pipelines, and standard fine-tuning baselines.
Finally, we provide ablation studies on training data efficiency and parameter scaling, followed by an out-of-domain evaluation to assess cross-task generalizability.
All training and experiments are run on a single workstation with an Intel Core i9-13900KF CPU, 128 GB memory, and one RTX 4090 GPU.
All source codes are available on GitHub\footnote{\raggedright Implementation: \url{https://github.com/VXRealLimited/DocMIDE}; benchmark: \url{https://github.com/VXRealLimited/DocMIDE-Benchmark}.}.

\input{sections/experiments/settings.tex}

\input{sections/experiments/main.tex}

\input{sections/experiments/ablation.tex}

\input{sections/experiments/others.tex}

\input{sections/experiments/size.tex}

%% file: sections/experiments/settings.tex
\subsection{Experimental Settings}
\label{sec:exp_settings}

\textbf{Datasets.} 
To evaluate multi-hop implicit derivation in document processing, we construct a dedicated benchmark dataset\footnote{\url{https://huggingface.co/datasets/VXRealLimited/DocMIDE}} comprising \textbf{4,151} document-field pairs.
This split includes 1,151 pairs sourced from real-world document images (extracted from UNIKIE-Bench \cite{ji2026unikie}) and 3,000 synthetically generated document images paired with derivative tasks to ensure scale and structural diversity.
For model optimization, we curate a training set of 100 annotated samples, utilizing a default subset of $n=50$ for primary model training and reserving the remainder to analyze training sample efficiency.
Furthermore, to evaluate whether \oursshort preserves general document understanding without catastrophic forgetting, we evaluate out-of-domain performance on OmniDocBench~\cite{ouyang2025omnidocbench} (scored on a 0--100 scale), ChartQAPro~\cite{masry2025chartqapro}, and the DocVQA validation set~\cite{mathew2021docvqa}.

\textbf{Baselines.}
We compare \oursshort against four primary categories of baseline approaches:
\begin{itemize}
    \item \textit{Traditional OCR pipelines:} Modular frameworks where an OCR engine first transcribes visual document text, which is subsequently processed by a downstream language model. Specifically, we evaluate GLM-OCR~\cite{duan2026glm} and BLOCKIE~\cite{bhattacharyya2025information}, both paired with a Qwen3.5-4B LLM backend.
    \item \textit{Base VLMs:} Direct multi-modal backbones evaluated off-the-shelf to measure zero-shot capabilities, including Qwen3.5-4B~\cite{qwen2026qwen35} and InternVL3.5-4B~\cite{wang2025internvl35}.
    \item \textit{Specialized VLMs:} Vision-language models customized for enhanced document comprehension, though lacking specific optimization for multi-hop implicit derivation, including Eagle2.5-8B~\cite{chen2026eagle} and Qianfan-OCR~\cite{dong2026qianfanocr}.
    \item \textit{Vision Reasoning Models:} Multimodal reasoning models optimized for open-ended, complex visual reasoning tasks (Vision-R1~\cite{huang2026vision} and StaR-KVQA~\cite{wen2026star}) evaluated to assess how general visual CoT capabilities transfer to schema-constrained implicit document extraction.
\end{itemize}

\textbf{Implementation Details.} 
We adopt Qwen3.5-4B~\cite{qwen2026qwen35} as our primary vision-language backbone due to its superior trade-off between deployment cost and baseline reasoning performance across vision-language tasks.
To evaluate parameter scalability and architectural generalizability, we also conduct experiments using smaller Qwen3.5 variants (2B and 0.8B) as well as InternVL3.5-4B \cite{wang2025internvl35}.
All models are trained under identical hyperparameter settings.

%% file: sections/experiments/main.tex
\subsection{Main Results}
\label{sec:main_results}


\begin{table}[!t]
\centering
\scriptsize
\setlength{\tabcolsep}{6pt}
\caption{Performance on multi-hop implicit derivation. Best performance is highlighted in \textbf{bold}.}
\label{tab:final}
\begin{tabular}{lccc}
\toprule
Model & Synthetic & Real & Overall \\
\midrule
\textbf{Traditional OCR Pipelines} & & & \\
\quad GLM-OCR~\cite{duan2026glm} & 32.9\% & 44.5\% & 36.1\% \\
\quad BLOCKIE~\cite{bhattacharyya2025information}  & 60.6\% & 68.6\% & 62.8\% \\
\midrule
\textbf{Base VLMs} & & & \\
\quad Qwen3.5-4B~\cite{qwen2026qwen35} & 65.1\% & 85.5\% & 70.8\% \\
\quad InternVL3.5-4B~\cite{wang2025internvl35} & 4.2\% & 37.6\% & 13.5\% \\
\midrule
\textbf{Specialized VLMs} & & & \\
\quad Eagle2.5-8B~\cite{chen2026eagle} & 0.0\% & 21.5\% & 6.0\% \\
\quad Qianfan-OCR~\cite{dong2026qianfanocr} & 34.4\% & 75.0\% & 45.7\% \\
\midrule
\textbf{Vision Reasoning Models} & & & \\
\quad Vision-R1~\cite{huang2026vision} & 71.3\% & 76.6\% & 72.8\% \\
\quad StaR-KVQA~\cite{wen2026star} & 93.9\% & 30.3\% & 76.3\% \\
\midrule
\textbf{Ours (\oursshort)} & & & \\
\quad Qwen3.5-4B & \textbf{98.8\%} & \textbf{88.2\%} & \textbf{95.9\%} \\
\quad InternVL3.5-4B & 85.5\% & 74.9\% & 82.6\% \\
\bottomrule
\end{tabular}
\end{table}

As shown in Table~\ref{tab:final}, \oursshort achieves state-of-the-art performance across both synthetic and real-world image distributions.
When applied to Qwen3.5-4B, our method reaches an overall accuracy of \textbf{95.9\%}, substantially outperforming all competing baselines.
Similarly, \oursshort elevates InternVL3.5-4B from 13.5\% to \textbf{82.6\%}, demonstrating that our framework generalizes across underlying vision-language backbones to unlock multi-hop implicit derivation.

Among non-adapted baselines, the Vision Reasoning Models are strongest apart from our method: StaR-KVQA reaches 76.3\% and Vision-R1 72.8\% overall, ahead of the best Base VLM, Qwen3.5-4B, at 70.8\%.
In contrast, Specialized VLMs yield lower performance—such as Qianfan-OCR at 45.7\% and Eagle2.5-8B at 6.0\%—because they are tailored primarily for visual text extraction and layout recognition rather than multi-hop implicit derivation.

Interestingly, some non-adapted baselines exhibit higher accuracy on real-world images than on synthetic ones (e.g., Qwen3.5-4B at 85.5\% real vs.\ 65.1\% synthetic).
This stems from real-world documents having fixed, constrained layouts where the implicit derivation complexity are inherently lower.
Conversely, \oursshort experiences a slight drop on real-world images (98.8\% synthetic to 88.2\% real), primarily due to degraded source image quality that induces text recognition errors prior to derivation.
Nevertheless, \oursshort (Qwen3.5-4B) still achieves the highest accuracy on real-world images (88.2\%).

%% file: sections/experiments/ablation.tex
\subsection{Effectiveness of Proposed Training Scheme}
\label{sec:training_efficacy}

\begin{table}[!t]
\centering
\scriptsize
\setlength{\tabcolsep}{6pt}
\caption{Held-out overall accuracy across Qwen3.5 model parameter scales (0.8B to 4B), comparing standard fine-tuning (FT) with \oursshort.}
\label{tab:scale}
\begin{tabular}{lccc}
\toprule
Model Size & Untuned Base & Standard FT & Ours (\oursshort) \\
\midrule
Qwen3.5-4B & 70.8\% & 69.0\% & 95.9\% \\
Qwen3.5-2B & 76.9\% & 54.0\% & 83.3\% \\
Qwen3.5-0.8B & 63.6\% & 25.3\% & 72.3\% \\
InternVL3.5-4B & 13.5\% & 54.2\% & 82.6\% \\
\bottomrule
\end{tabular}
\end{table}

In Table~\ref{tab:scale}, we evaluate the effectiveness of our training method against Standard FT across model parameter scales and architectures.
As analyzed in Section~\ref{sec:naive}, Standard FT is often ineffective and can even severely degrade baseline performance—most notably on Qwen3.5-0.8B, where overall accuracy drops precipitously from 63.6\% down to 25.3\%.

This degradation occurs because Standard FT provides target demonstrations without granular reward feedback on intermediate, partially correct reasoning steps.
Qualitative inspection reveals that while Standard FT models attempt CoT reasoning, they fail to break complex logic into atomic operations.
For example, a Standard FT Qwen3.5-0.8B instance outputs: \textit{``Scanning the row totals, the Espresso (\$2.50), Croissant (\$3.25), Blueberry Muffin (\$3.75), Iced Latte (\$4.50), Oat Milk Carton (\$1.25), and Chocolate Slice (\$5.00) all add up to \$20.00 directly.''}
Here, the model correctly identifies visual elements but attempts multi-item arithmetic in a single leap, resulting in calculation failure.

In contrast, \oursshort leverages explicit reward signals to guide the model through discrete, atomic derivation steps.
By reinforcing each correct sub-step, \oursshort prevents multi-step reasoning collapse and yields consistent performance improvements across all model scales, boosting Qwen3.5-0.8B to 72.3\% and Qwen3.5-4B to 95.9\%.

%% file: sections/experiments/others.tex
\subsection{Trade-Off Analysis on Out-of-Domain Tasks}
\label{sec:ood_tradeoff}

\begin{table}[!t]
\centering
\scriptsize
\setlength{\tabcolsep}{6pt}
\caption{Evaluation on out-of-domain benchmarks: OmniDocBench~\cite{ouyang2025omnidocbench} (0--100 scale), ChartQAPro~\cite{masry2025chartqapro}, and DocVQA~\cite{mathew2021docvqa} validation set. Best performance in each column is highlighted in \textbf{bold}.}
\label{tab:ood}
\begin{tabular}{lccc}
\toprule
& Page Parsing & \multicolumn{2}{c}{Question Answering} \\
\cmidrule(lr){2-2} \cmidrule(lr){3-4}
Model & OmniDocBench & ChartQAPro & DocVQA \\
\midrule
\textbf{Traditional OCR Pipelines} & & & \\
\quad GLM-OCR~\cite{duan2026glm} & 60.1 & -- & -- \\
\quad BLOCKIE~\cite{bhattacharyya2025information}  & 44.3 & 42.0\% & 48.6\% \\
\midrule
\textbf{Base VLMs} & & & \\
\quad Qwen3.5-4B~\cite{qwen2026qwen35} & 61.8 & 15.8\% & 81.9\% \\
\quad InternVL3.5-4B~\cite{wang2025internvl35} & 76.2 & 30.2\% & 76.9\% \\
\midrule
\textbf{Specialized VLMs} & & & \\
\quad Eagle2.5-8B~\cite{chen2026eagle} & 6.2 & 22.4\% & 28.2\% \\
\quad Qianfan-OCR~\cite{dong2026qianfanocr} & \textbf{85.8} & 38.5\% & 80.6\% \\
\midrule
\textbf{Vision Reasoning Models} & & & \\
\quad Vision-R1~\cite{huang2026vision} & 13.0 & 34.0\% & 91.7\% \\
\quad StaR-KVQA~\cite{wen2026star} & 77.2 & 38.9\% & 76.3\% \\
\midrule
\textbf{Ours (\oursshort)} & & & \\
\quad Qwen3.5-4B & 77.1 & \textbf{50.5\%} & \textbf{92.7\%} \\
\quad InternVL3.5-4B & 70.2 & 26.6\% & 78.9\% \\
\bottomrule
\end{tabular}
\end{table}

To verify whether specialized training incurs a trade-off on broader vision-language capabilities, we evaluate models on three unseen out-of-domain benchmarks (Table~\ref{tab:ood}).
While out-of-domain optimization is not the primary focus of our work, this audit ensures that targeted derivation alignment does not induce severe catastrophic forgetting.
For Qwen3.5-4B, \oursshort not only avoids capability degradation but actually yields substantial positive transfer across general tasks, notably boosting performance on OmniDocBench (from 61.8 to 77.1), ChartQAPro (from 15.8\% to 50.5\%), and DocVQA (from 81.9\% to 92.7\%) --- outperforming even the dedicated vision reasoning models on both question-answering benchmarks.
When applied to InternVL3.5-4B, \oursshort exhibits minimal trade-off: it retains strong performance on OmniDocBench (70.2 vs.\ 76.2 base) and ChartQAPro (26.6\% vs.\ 30.2\% base), while slightly improving on DocVQA (78.9\% vs.\ 76.9\% base).

%% file: sections/experiments/size.tex
\subsection{Training Sample Efficiency}

\begin{figure}[!t]
\centering
\includegraphics[width=\linewidth]{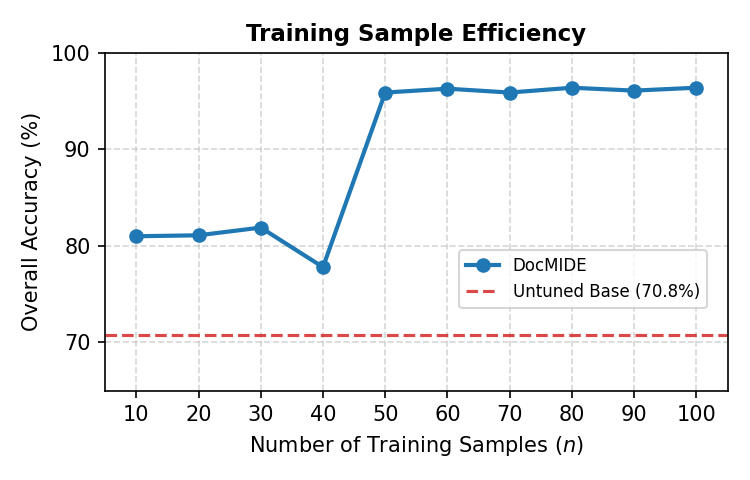}
\caption{Held-out overall accuracy across varying numbers of annotated training samples $n$ using \oursshort on Qwen3.5-4B.}
\label{fig:data_scaling}
\end{figure}


To determine minimum data requirements, we evaluate the sample efficiency of \oursshort by scaling the number of annotated training samples $n$ from 10 to 100 on Qwen3.5-4B (Figure~\ref{fig:data_scaling}).
The results show that fine-tuning with as few as $n=10$ samples raises overall accuracy from 70.8\% to 81.0\%, demonstrating immediate efficiency.
Performance increases sharply up to $n=50$, where it achieves 95.9\% accuracy, after which gains plateau ($96.4\%$ at $n=100$).

These results confirm that the manual annotation effort required by our framework is remarkably lightweight.
Requiring only 50 labeled examples to reach peak performance makes our method easily adaptable to custom document domains without extensive data collection.

%% file: sections/limitations.tex
\section{Conclusion}

In this paper, we presented \oursshort, a lightweight fine-tuning framework that solves multi-hop implicit derivation in visually rich documents.
By combining GRPO with a deterministic, four-component reward function, DocMIDE trains compact vision-language models ($\le$4B parameters) to explicitly retrieve visual evidence before computing target values.
Experiments show that DocMIDE significantly outperforms base VLMs, OCR pipelines, and standard fine-tuning methods while maintaining high sample efficiency.
Overall, DocMIDE provides an effective foundation for verifiable visual reasoning, laying the groundwork for extensions to multi-page documents and larger multimodal systems.

%% file: sections/appendix.tex
\section{Is the Standard Fine-Tuning Baseline Undertrained?}
\label{sec:appendix_sft_scale}

The standard fine-tuning baseline reported in Table~\ref{tab:scale} is trained on the same $50$ exemplars as \oursshort, so that the two differ only in training scheme and not in supervision budget.
This raises an obvious objection: the baseline may simply be undertrained, and its failure on multi-hop derivation may reflect a shortage of demonstrations rather than a limitation of demonstration-only supervision.
To test this, we fine-tune Qwen3.5-4B on the full pool of $200$ verified exemplars under an otherwise identical recipe and evaluate on the same 4{,}151-pair benchmark.

\begin{table}[!h]
\centering
\scriptsize
\setlength{\tabcolsep}{6pt}
\caption{Standard fine-tuning of Qwen3.5-4B at two supervision budgets, evaluated on the same 4{,}151-pair benchmark. Quadrupling the number of verified exemplars does not recover multi-hop derivation.}
\label{tab:sft200}
\begin{tabular}{lccc}
\toprule
Training exemplars & Synthetic & Real & Overall \\
\midrule
50 (used in Table~\ref{tab:scale}) & 63.6\% & 83.2\% & 69.0\% \\
200 (full pool) & 62.9\% & 83.8\% & 68.7\% \\
\bottomrule
\end{tabular}
\end{table}

Table~\ref{tab:sft200} shows that quadrupling the supervision changes almost nothing.
Overall accuracy moves from 69.0\% to 68.7\%.
The two splits behave the same way: synthetic accuracy is flat (63.6\% to 62.9\%) and real accuracy shifts by half a point (83.2\% to 83.8\%).
Every one of these gaps is smaller than the run-to-run spread visible in the sample-efficiency sweep of Figure~\ref{fig:data_scaling}, so we read the curve as flat rather than as evidence of a small gain or loss.

The failure is also unchanged in kind, not merely in magnitude.
Under the $200$-exemplar budget the model still collapses specifically on the derivations that require the longest operator chains: accuracy on the long-summation subset of the synthetic split is $26.3\%$, against $92.5\%$ on the category-aggregation subset, whose chains are short.
This is the step-skipping mode of Section~\ref{sec:naive} reproduced at four times the data: the model reads the operands correctly and then asserts a long total in a single leap.
Additional demonstrations of the correct trace format do not teach the model to execute the chain, because nothing in the supervised objective distinguishes a trace that carries out every intermediate step from one that skips to a plausible endpoint.
Rewarding the intermediate steps directly, as in Section~\ref{sec:reward}, is what closes the gap.